\documentclass{article}
\usepackage[T1]{fontenc}
\usepackage{iclr2027_conference,times}
\usepackage{amsmath,amssymb,mathtools}
\usepackage{graphicx,booktabs,array,multirow}
\usepackage{hyperref}
\usepackage{url}
\usepackage{bbm}
\usepackage{booktabs}
\usepackage{float}
\usepackage{enumitem}
\usepackage{xspace}
\usepackage{xcolor}
\usepackage{pifont}

\iclrfinalcopy

\newcommand{\cmark}{\textcolor{green!60!black}{\ding{51}}}
\newcommand{\xmark}{\textcolor{red}{\ding{55}}}

\usepackage{array} 
\usepackage{threeparttable}
\usepackage{algorithm}
\usepackage{algorithmic}
\usepackage{xspace}

\newcommand{\ourmethod}{DynStream}
\hypersetup{colorlinks=true,linkcolor=blue!45!black,citecolor=blue!45!black,urlcolor=blue!45!black,pdftitle={\ourmethod{}: Online Feed-forward Reconstruction of Dynamic Worlds from Unposed Video},pdfauthor={}}
\title{\ourmethod{}: Online Streaming 4D Gaussian Reconstruction of Dynamic Worlds from Unposed Video}

\author{
Dingwei Xian\textsuperscript{1}$^\dagger$
~ Xiaoyu Zhou\textsuperscript{1}$^\dagger$
~ Yajiao Xiong\textsuperscript{1}
~ \textbf{Yongtao Wang\textsuperscript{1,2}}\thanks{Corresponding author.}
~ \textbf{Ming-Hsuan Yang\textsuperscript{3}} \\
{\textsuperscript{1}Wangxuan Institute of Computer Technology, Peking University} \\
~ {\textsuperscript{2}VGI Labs Co., Ltd.} 
~ {\textsuperscript{3}University of California, Merced} \\
}

\begin{document}
\maketitle

\lhead{Preprint}

{
\begin{figure}[h]
\hsize=\textwidth
\centering
\includegraphics[width=1.0\linewidth]{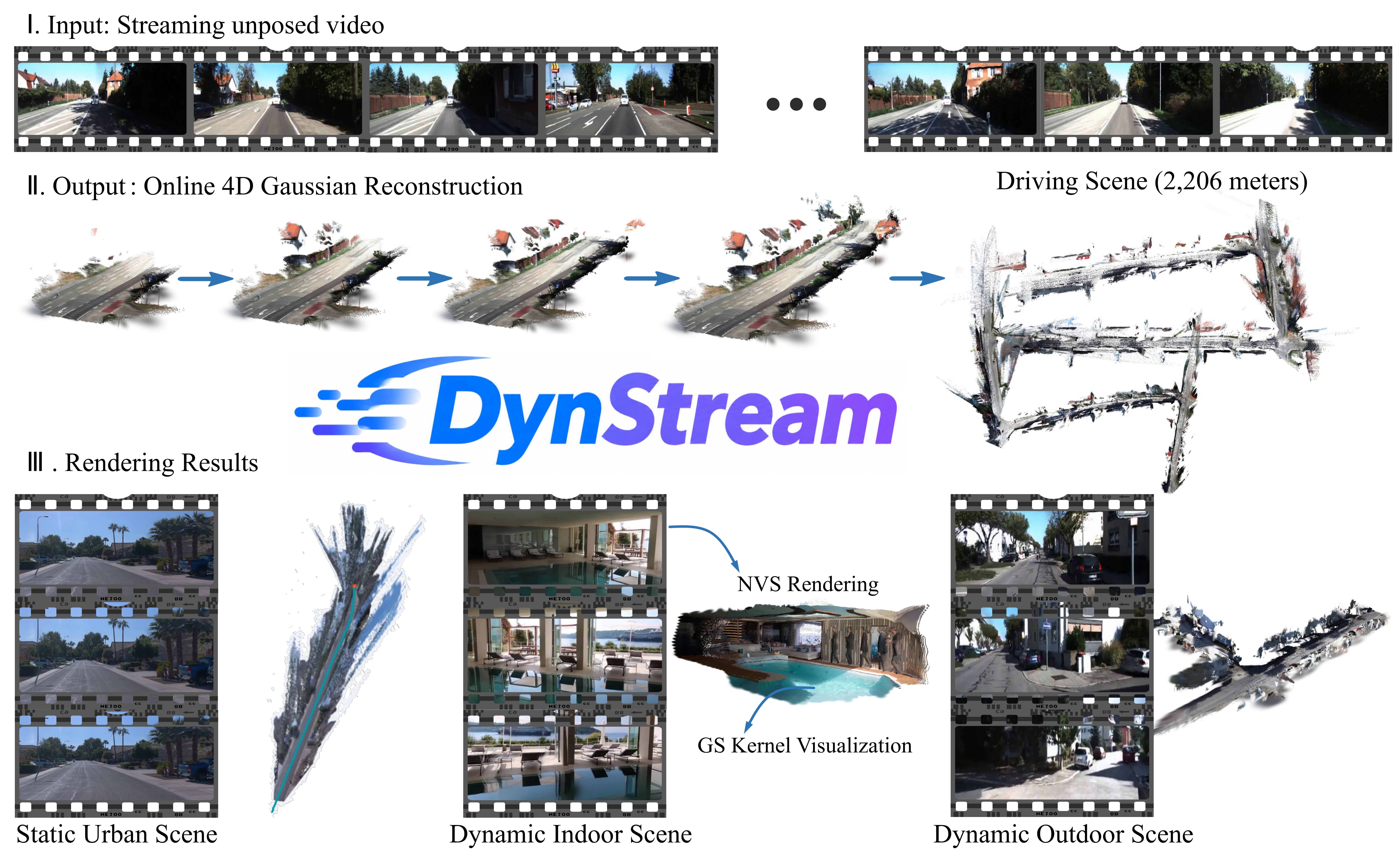}
\vspace{-10pt}
\caption{\textbf{\ourmethod{}} enables online, photorealistic 4D reconstruction from long, unposed video streams without per-scene optimization. It incrementally integrates incoming observations into a globally consistent representation, supports arbitrarily long streaming inputs, and generalizes across diverse dynamic and static indoor/outdoor scenes.
}
\label{fig:teaser}
\end{figure}
}

\renewcommand{\thefootnote}{}
\footnotetext{$^\dagger$ denotes equal contribution.}

\begin{abstract}
Online reconstruction of dynamic 4D scenes from long, unposed streaming videos requires both continuous processing and photorealistic rendering, which existing methods struggle to achieve simultaneously. Existing feed-forward Gaussian methods are restricted to offline processing, whereas online point-cloud approaches struggle to maintain dense geometry and high-fidelity rendering.

We present \textbf{\ourmethod{}}, a framework for streaming 4D Gaussian reconstruction from long, unposed videos. Given a continuous video stream, \ourmethod{} reconstructs the scene within local temporal windows and incrementally aligns and fuses these local reconstructions into a globally consistent scene, enabling online 4D reconstruction without per-scene optimization. By jointly enforcing cross-window geometric consistency and modeling time-varying scene content, \ourmethod{} supports efficient reconstruction and photorealistic rendering over extended video streams.

Experiments demonstrate that \ourmethod{} enables high-fidelity online dynamic reconstruction and rendering from long video streams, achieving state-of-the-art performance across diverse dynamic indoor and outdoor scenes.

\end{abstract}

\section{Introduction}
\label{sec:intro}

Reconstructing dynamic 4D worlds from continuous video streams is a fundamental step toward building persistent, photorealistic representations of real-world environments. Unlike curated image sequences or short video clips, real-world streams are typically long and unposed, with unconstrained camera motion and continuously changing scene content. A practical reconstruction system must therefore incrementally integrate observations over time while preserving both geometric consistency and high-fidelity rendering. Such capability is essential for applications including autonomous driving, embodied intelligence, and interactive content creation. However, achieving continuous 4D reconstruction remains challenging, as the system must simultaneously maintain long-term temporal consistency, capture dynamic scene evolution, and operate efficiently under online processing constraints.

Existing approaches to online dynamic scene reconstruction largely follow two paradigms, both of which face fundamental limitations for long, continuously arriving video streams. Optimization-based methods \citep{newcombe2015dynamicfusion,slavcheva2017killingfusion,yu2018doublefusion} incrementally update scene representations as new observations arrive, including recent Gaussian-based approaches \citep{li2026dynagslam} for photorealistic online mapping. However, their reliance on scene-specific optimization and iterative tracking causes computational costs to grow with sequence length and can lead to accumulated tracking and reconstruction errors over time. Many methods also rely on depth estimates or motion priors, further limiting their applicability to unconstrained in-the-wild videos. Fundamentally, these approaches lack a scalable mechanism for incrementally incorporating long-term observations without repeatedly optimizing the scene.

Recent feed-forward methods \citep{ye2025noposplat,hong2025pf3plat,jiang2025anysplat,ye2026yonosplat,yang2025storm,yang2026neoverse} avoid costly per-scene optimization by directly predicting scene representations from visual observations. However, they typically assume a fixed or bounded set of frames processed jointly in a single forward pass and operate on pre-recorded video clips, preventing newly arriving observations from being incorporated without reprocessing previous frames. As a result, computation and memory grow with sequence length, making them poorly suited to long video streams. Recent online feed-forward methods based on point clouds \citep{wang2024dust3r,zhang2025monst3r,wang2025vggt} extend this paradigm to streaming reconstruction, but typically require ground-truth 3D geometry and camera poses for training, limiting scalable self-supervised learning and making joint estimation of camera motion and scene geometry difficult. Moreover, these methods often fail to simultaneously provide dense geometric predictions and photorealistic rendering. Thus, existing feed-forward approaches remain ill-suited to high-quality online reconstruction from long, unposed video streams due to their bounded temporal context, strong supervision requirements, or limited reconstruction fidelity.

To overcome these limitations, we present \textbf{\ourmethod{}}, a framework for online reconstruction of long, unposed video streams into globally consistent 4D Gaussian scenes, jointly estimating camera motion and time-varying scene representations while enabling photorealistic rendering without per-scene optimization or 3D/pose supervision.

Our method is built on two key components. First, a \emph{causal-attention online reconstruction module} incrementally processes incoming observations and predicts camera poses and newly observed Gaussian primitives using only past and current frames. Second, a \emph{dynamic-aware relative-pose strategy} establishes spatial consistency across successive observations by accounting for time-varying scene content during pose estimation and scene integration.

Specifically, we divide the incoming stream into consecutive temporal windows, or \emph{chunks}. Within each chunk, the first frame serves as a keyframe, and subsequent camera poses are predicted relative to it along with the corresponding Gaussian representation. Because independently reconstructed chunks may reside in different coordinate gauges, particularly in scale, we additionally predict a metric scale factor to align each local reconstruction with a shared metric coordinate system. By incrementally processing and integrating successive chunks, \ourmethod{} enables online reconstruction of long dynamic video streams while preserving globally consistent geometry and high-fidelity scene representations.

The main contributions of this work are:

\begin{itemize}[leftmargin=*,itemsep=1pt,topsep=1pt,partopsep=0pt]
\item We introduce \ourmethod{}, a framework for online dynamic 4D Gaussian reconstruction from long, unposed video streams, enabling continuous scene construction and photorealistic rendering through incremental integration of incoming observations.
\item We develop a dynamic-aware relative-pose strategy that propagates camera and scene coordinates across successive chunks, enabling long-sequence dynamic reconstruction without independent chunk registration or per-scene optimization.
\item
Experiments demonstrate that \ourmethod{} achieves competitive rendering quality while enabling online dynamic scene reconstruction from long video streams, supporting efficient and scalable online 4D reconstruction across diverse scenes.
\end{itemize}

\section{Related Work}
\label{sec:related}

\paragraph{Online Dynamic Scene Reconstruction}

Online reconstruction of dynamic scenes has traditionally been formulated as an iterative tracking-and-mapping problem, where an evolving scene representation is continuously updated as new observations arrive
\citep{newcombe2015dynamicfusion,slavcheva2017killingfusion,yu2018doublefusion}.
These approaches typically couple scene integration with correspondence estimation, motion tracking, and iterative optimization to maintain geometric consistency over time
\citep{newcombe2015dynamicfusion,slavcheva2017killingfusion,yu2018doublefusion},
and often rely on explicit depth observations to support robust online reconstruction.
More recent Gaussian-based methods extend this paradigm with directly renderable scene representations, jointly modeling scene geometry, camera motion, and dynamic content to enable photorealistic online reconstruction
\citep{li2026dynagslam}.
Nevertheless, these approaches remain fundamentally optimization- or tracking-driven, requiring repeated scene-specific computation as new observations arrive
\citep{li2026dynagslam}.
In contrast, our goal is to retain the incremental nature of online reconstruction while replacing repeated scene-specific optimization with direct feed-forward inference from RGB observations.

\paragraph{Feed-Forward Reconstruction}

Recent feed-forward reconstruction methods shift substantial geometric reasoning from test-time optimization into learned inference, directly estimating camera poses, depth, point-based geometry, and motion from visual observations
\citep{wang2024dust3r,zhang2025monst3r,wang2025vggt}.
Building on this paradigm, recent works further introduce causal attention, temporal feature propagation, keyframe-relative pose estimation, and metric scale prediction to extend feed-forward geometry reconstruction toward long and streaming sequences
\citep{zhuo2026streamvggt,cheng2026longstream}.
These approaches demonstrate that geometric reconstruction can be performed efficiently without repeatedly optimizing over the entire observation history. However, their outputs primarily describe geometric quantities rather than a continuously maintained, photorealistic scene representation that can be directly rendered and incrementally updated.

Feed-forward Gaussian reconstruction addresses this limitation by directly predicting renderable Gaussian primitives, together with camera parameters when necessary, from posed or unposed visual observations
\citep{ye2025noposplat,hong2025pf3plat,jiang2025anysplat,ye2026yonosplat}.
More recent approaches extend this formulation to dynamic scenes by jointly modeling time-varying geometry, appearance, motion, and camera poses
\citep{yang2025storm,chen2026dggt,yang2026neoverse},
while streaming formulations have begun to process dynamic observations causally with bounded temporal context \citep{qiu2026slarm}.

Despite this progress, existing methods either operate on pre-collected image sets or video clips, or require known camera parameters for streaming reconstruction, and thus do not continuously build a globally aligned, renderable scene from long, unposed video streams. In contrast, \ourmethod{} combines causal feed-forward reconstruction with direct Gaussian prediction, incrementally propagating camera and scene coordinates across temporal chunks and integrating them into a shared representation without reprocessing past frames or per-scene optimization.

\section{Method}
\label{sec:method}
\subsection{Problem Formulation}
\label{sec:problem_setup}

Given a long unposed video stream $\mathcal{V}_{1:T} = \{I_i\}_{i=1}^{T}$, our goal is to incrementally reconstruct a dynamic scene as a collection of Gaussians while simultaneously estimating the camera trajectory. Formally, our model implements the mapping:
\begin{equation}
\label{eq:problem_mapping}
F_{\theta}:\quad
\mathcal{V}_{1:t}
\longmapsto
\mathcal{S}_t
=
(\mathcal{G}_t,\mathcal{T}_{1:t}),
\end{equation}
where the output $\mathcal{S}_t$ contains a time-aware scene representation and the corresponding camera trajectory. Specifically, $\mathcal{G}_t$ denotes the Gaussian scene representation maintained at time $t$, while $\mathcal{T}_{1:t}=(\mathbf{T}_1^{w},\ldots,\mathbf{T}_t^{w})$ denotes the estimated camera trajectory, with $\mathbf{T}_i^{w}\in\mathrm{SE}(3)$ representing the camera-to-world pose of frame $I_i$.

However, directly processing the entire growing prefix $\mathcal{V}_{1:t}$ with any ViT-based architecture is computationally expensive. With $P$ tokens per frame, attention over $t$ frames involves $O(t^2P^2)$ pairwise interactions and quadratic memory consumption \citep{dao2022flashattention,dosovitskiy2021vit}. We therefore divide the incoming stream into fixed-length consecutive temporal windows, referred to as \textbf{chunks}. Each chunk $\mathcal{C}_k=(I_{a_k},\ldots,I_{b_k})$ contains a bounded number of consecutive frames, and its first frame $I_{a_k}$ serves as the keyframe. During online processing, only the frames that have already arrived within the current chunk are visible to the model.

After organizing the stream into chunks, we can realize the mapping $F_{\theta}$ in Eq.~\eqref{eq:problem_mapping} through two sequential stages: causal local reconstruction and global scene integration:
\begin{equation}
\label{eq:online_mapping}
\mathcal{S}_t
=
F_{\theta}(\mathcal{V}_{1:t})
=
\Psi
\left(
\mathcal{S}_{t-1},
\Phi_{\theta}(\mathcal{C}_{k,\leq t},\mathcal{H}_{k-1})
\right).
\end{equation}
In this formulation, $\Phi_{\theta}$ reconstructs the newly observed content from the visible portion of the current chunk $\mathcal{C}_{k,\leq t}$ and the retained reference information $\mathcal{H}_{k-1}$, while $\Psi$ aligns the local prediction with the existing global scene and updates the scene state $\mathcal{S}_{t-1}$. We initialize $\mathcal{H}_0$ as empty and $\mathcal{S}_0$ with an empty scene and trajectory.

This formulation allows the model to process only a bounded temporal context at each step while preserving a coherent scene representation over long video streams. We implement $\Phi_{\theta}$ with a causal-attention online reconstruction module and $\Psi$ with a dynamic-aware relative-pose strategy, which are detailed in Sections~\ref{sec:causal_reconstruction} and~\ref{sec:dynamic_alignment} respectively.

\begin{figure*}[t]
\centering

\includegraphics[width=\textwidth]{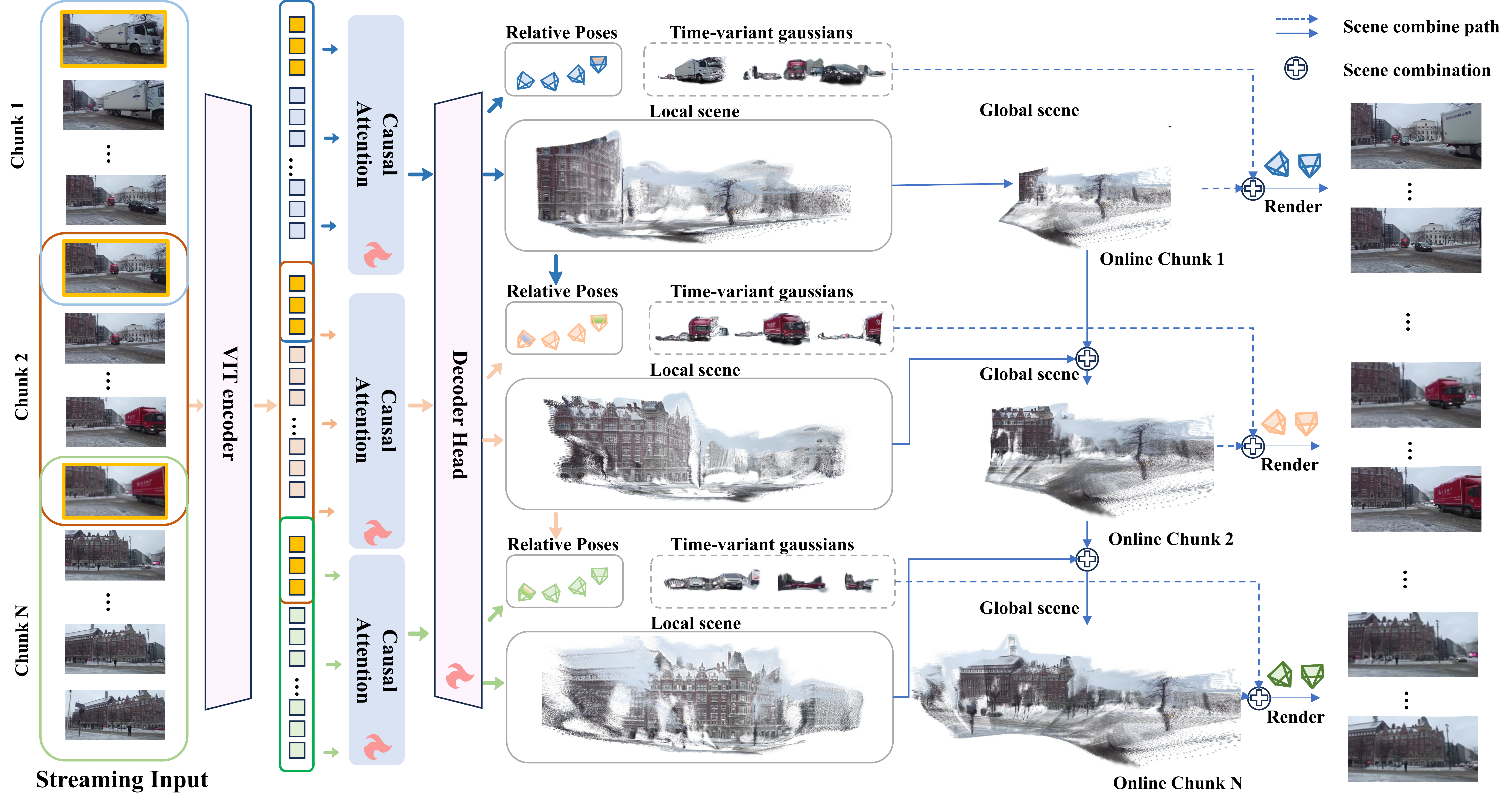}

\caption{
\textbf{Overview of \ourmethod{}.} The incoming video stream is processed sequentially in fixed-length chunks. For each chunk, the causal stream backbone predicts keyframe-relative camera poses, a local Gaussian scene, and time-varying Gaussian primitives. The local scene is aligned to the global coordinate system and incrementally fused with the previously reconstructed scene, forming a persistent global representation. Time-varying Gaussians preserve their temporal associations and are incorporated into the global scene only at the corresponding time. This design enables continuous reconstruction and rendering of long dynamic video streams while maintaining globally consistent static geometry and temporally coherent dynamic content.
}
\end{figure*}
\vspace{-6pt}

\subsection{Causal-Attention Online Reconstruction}
\label{sec:causal_reconstruction}

Following Eq.~\eqref{eq:online_mapping}, the local reconstruction operator $\Phi_{\theta}$ converts the currently available observations into a keyframe-relative reconstruction. We implement $\Phi_{\theta}$ with a DINOv2 image encoder \citep{oquab2024dinov2}, a causal alternating-attention backbone, and task-specific prediction heads.

\paragraph{Causal alternating-attention backbone.}
For each input frame $I_i\in\mathbb{R}^{H\times W\times 3}$, we divide the image into non-overlapping patches and embed them into visual tokens using DINOv2 \citep{oquab2024dinov2}. Following VGGT \citep{wang2025vggt}, the tokens are processed by alternating intra-frame and cross-frame attention blocks. Intra-frame attention aggregates spatial information within each image, while cross-frame attention exchanges information across observations. To satisfy the online setting in Eq.~\eqref{eq:problem_mapping}, we apply a causal mask so that frame $I_t$ can only attend to observations available up to time $t$.

For the visible prefix of the current chunk, the backbone produces:
\begin{equation}
\label{eq:causal_feature}
\mathbf{F}_{k,t}
=
A_{\theta}
\left(
E_{\theta}(\mathcal{C}_{k,\leq t})
\right),
\end{equation}
where $E_{\theta}$ denotes the DINOv2 patch encoder and $A_{\theta}$ denotes the causal alternating-attention. This allows $\Phi_{\theta}$ to aggregate spatial and temporal context without accessing future observations.

\paragraph{Multi-head local reconstruction.}
Task-specific heads decode $\mathbf{F}_{k,t}$ into the predictions required for local reconstruction:
\begin{equation}
\label{eq:local_prediction}
(
\mathbf{T}_{k,t},
D_{k,t},
G_{k,t},
s_{k,t}
)
=
\Phi_{\theta}
(
\mathcal{C}_{k,\leq t},
\mathcal{H}_{k-1}
).
\end{equation}
Here, $\mathbf{T}_{k,t}\in\mathrm{SE}(3)$ denotes the camera pose of frame $I_t$ relative to the current keyframe, distinguishing it from the global camera-to-world pose $\mathbf{T}_t^{w}$ in Eq.~\eqref{eq:problem_mapping}. The depth head predicts $D_{k,t}\in\mathbb{R}^{H\times W\times 1}$, while the Gaussian head predicts a pixel-aligned Gaussian map $G_{k,t}\in\mathbb{R}^{H\times W\times C_G}$, with one Gaussian primitive associated with each spatial location.

Using RGB appearance, we set $C_G=15$ and parameterize each Gaussian as:
\begin{equation}
\label{eq:gaussian_parameterization}
\mathbf{g}_{k,t}(u,v)
=
[
\boldsymbol{\mu},
\mathbf{q},
\mathbf{s},
\alpha,
\mathbf{c},
p_{\mathrm{dyn}}
]
\in
\mathbb{R}^{15},
\end{equation}
where $\boldsymbol{\mu}\in\mathbb{R}^{3}$ is the Gaussian center, $\mathbf{q}\in\mathbb{R}^{4}$ its rotation quaternion, $\mathbf{s}\in\mathbb{R}^{3}$ its anisotropic scale, $\alpha\in\mathbb{R}$ its opacity, $\mathbf{c}\in\mathbb{R}^{3}$ its RGB appearance, and $p_{\mathrm{dyn}}\in[0,1]$ its dynamic probability.

Although the predicted pose, depth, and Gaussian geometry share the same local coordinate system, independently reconstructed chunks can still differ by an unknown scale. This scale ambiguity affects camera translation, depth, Gaussian centers, and Gaussian sizes, preventing direct cross-chunk integration. We therefore introduce a scale head that predicts a positive metric factor $s_{k,t}>0$, which is applied consistently to all length-related predictions before global alignment.

The resulting local reconstruction is then passed to $\Psi$, which aligns it with the global coordinate system and updates the maintained scene representation.

\subsection{Relative-Pose Global Alignment and Dynamic-Aware Integration}
\label{sec:dynamic_alignment}

Following Eq.~\eqref{eq:online_mapping}, the alignment operator $\Psi$ transforms the local reconstruction produced by $\Phi_{\theta}$ into the shared world coordinate system and updates the persistent scene state. This process consists of relative-pose propagation, metric Gaussian alignment, and dynamic-aware scene integration.

\paragraph{Relative-pose global alignment.}
From Eq.~\eqref{eq:local_prediction}, $\mathbf{T}_{k,t}$ represents the pose of frame $I_t$ relative to the keyframe of chunk $k$. We write $\mathbf{T}_{k,t}=(\mathbf{R}_{k,t},\mathbf{u}_{k,t})$, where $\mathbf{R}_{k,t}$ and $\mathbf{u}_{k,t}$ denote the relative rotation and translation, respectively.

Let $\mathbf{K}_k\in\mathrm{SE}(3)$ denote the camera-to-world pose of the keyframe of chunk $k$. At each chunk transition, the relative-pose head predicts a transformation $\mathbf{L}_k$ between the new keyframe and the previous keyframe using the retained reference information $\mathcal{H}_{k-1}$. Starting from $\mathbf{K}_1=\mathbf{I}_4$, the global keyframe pose and the global pose of frame $I_t$ are obtained as:
\begin{equation}
\label{eq:global_pose}
\mathbf{K}_k
=
\mathbf{K}_{k-1}\mathbf{L}_k,
\qquad
\mathbf{T}_t^{w}
=
\mathbf{K}_k
\begin{bmatrix}
\mathbf{R}_{k,t} & s_{k,t}\mathbf{u}_{k,t} \\
\mathbf{0}^{\mathsf{T}} & 1
\end{bmatrix}.
\end{equation}
The predicted scale $s_{k,t}$ converts the local translation into metric units. The same calibration is applied to the predicted depth as $D_{k,t}^{\mathrm{m}}=s_{k,t}D_{k,t}$. In this way, local camera predictions from different chunks are progressively connected to the same global coordinate system without global trajectory optimization.

\paragraph{Metric Gaussian alignment.}
We apply the same scale and coordinate transformation to the local Gaussian map $G_{k,t}$. We denote this operation by:
\begin{equation}
\label{eq:gaussian_alignment}
G_{k,t}^{w}
=
\mathcal{W}
\left(
G_{k,t};
\mathbf{K}_k,
s_{k,t}
\right),
\end{equation}
where $\mathcal{W}$ transforms the local Gaussian parameters into the world coordinate system. Specifically, Gaussian centers and spatial scales are multiplied by $s_{k,t}$, while their positions and orientations are transformed according to $\mathbf{K}_k$. Appearance, opacity, and dynamic probability remain unchanged. As a result, $G_{k,t}^{w}$ can be directly integrated with Gaussian predicted from previous chunks.

\paragraph{Dynamic-aware scene integration.}
Although the Gaussian primitives are spatially aligned, dynamic content observed at different times should not be indiscriminately accumulated into the same persistent representation. We therefore use the dynamic probability $p_{\mathrm{dyn}}$ contained in each Gaussian to distinguish persistent and time-varying content. Given a threshold $\tau$, Gaussians with $p_{\mathrm{dyn}}\leq\tau$ are treated as static and integrated into the persistent scene, while Gaussians with $p_{\mathrm{dyn}}>\tau$ are treated as dynamic and retain their temporal association with the corresponding observation.

The alignment operator $\Psi$ therefore updates the scene state by jointly incorporating the globally aligned Gaussian map $G_{k,t}^{w}$ and the camera pose $\mathbf{T}_t^{w}$. Static structure is progressively accumulated across chunks, whereas dynamic content remains time-specific. This allows successive local reconstructions produced by $\Phi_{\theta}$ to form a spatially consistent and temporally aware global scene.

\begin{figure*}[ht]
\centering

\includegraphics[width=\textwidth]{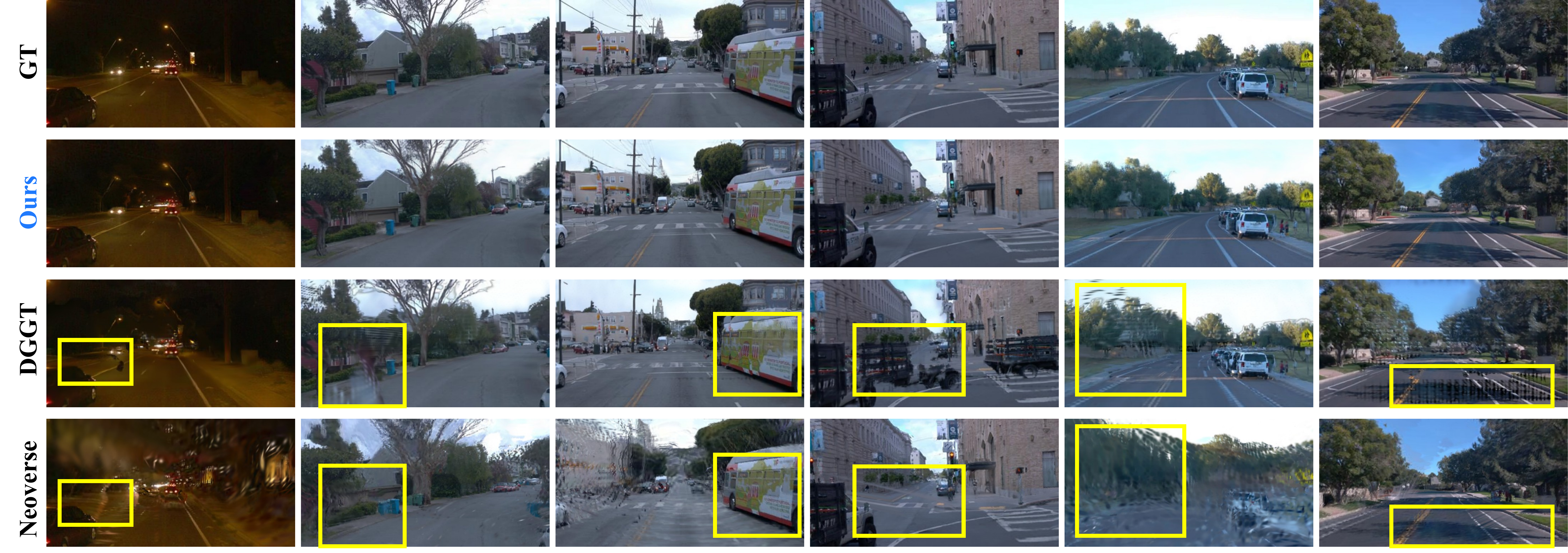}

\caption{Qualitative comparison of NVS rendering results on the Waymo Open Dataset. Each column shows one scene, and each row corresponds to GT, \textbf{Ours}, DGGT, and NeoVerse, respectively.}
\label{fig:waymo_qualitative}

\end{figure*}

\subsection{Loss Function}
\label{sec:loss}

We employ a frozen VGGT-$\Omega$~\citep{wang2026vggtomega} teacher to provide pseudo geometric and camera supervision during training. Based on these teacher predictions and the available metric and dynamic annotations, we train the model with five objectives that supervise appearance, geometry, camera motion, metric scale, and dynamic scene decomposition:
\begin{equation}
\label{eq:total_loss}
\mathcal{L}
=
\mathcal{L}_{\mathrm{render}}
+
\lambda_{1}\mathcal{L}_{\mathrm{depth}}
+
\lambda_{2}\mathcal{L}_{\mathrm{pose}}
+
\lambda_{3}\mathcal{L}_{\mathrm{scale}}
+
\lambda_{4}\mathcal{L}_{\mathrm{dyn}}.
\end{equation}

\paragraph{Online rendering loss.}
To preserve the online setting during training, rendering supervision is applied to the causal scene state $\mathcal{G}_t$, which contains only Gaussian primitives reconstructed from observations available up to time $t$. We render the current observation using its estimated global camera pose:
\begin{equation}
(\widetilde{I}_t,\widetilde{D}_t)
=
\mathcal{R}(\mathcal{G}_t,\mathbf{T}_t^{w}),
\end{equation}
where $\mathcal{R}$ denotes the differentiable Gaussian rasterizer, and $\widetilde{I}_t$ and $\widetilde{D}_t$ are the rendered RGB image and depth map, respectively. Following~\citep{jiang2025anysplat}, we supervise rendering using a combination of pixel-wise reconstruction and perceptual losses:
\begin{equation}
\mathcal{L}_{\mathrm{render}}
=
\|\widetilde{I}_t-I_t\|_2^2
+
\lambda_{5}
\operatorname{LPIPS}(\widetilde{I}_t,I_t).
\end{equation}

\paragraph{Depth loss.}
Following AnySplat~\citep{jiang2025anysplat}, we supervise the depth branch using pseudo ground-truth depth $D_{\mathrm{gt}}$ predicted by the teacher model, together with a depth consistency term that enforces consistency between the predicted depth and the Gaussian representation:
\begin{equation}
\label{eq:depth_loss}
\mathcal{L}_{\mathrm{depth}}
=
\|\hat{D}_{t}-D_{\mathrm{gt}}\|_2^2
+
\lambda_{6}
\|\hat{D}_t-\widetilde{D}_{t}\|_2^2.
\end{equation}

\paragraph{Relative pose loss.}
We supervise camera motion in the keyframe-relative gauge. For frame $I_t$, the teacher relative pose is obtained from the teacher camera poses as $(\mathbf{T}_{r_k}^{\mathrm{tea}})^{-1}\mathbf{T}_{t}^{\mathrm{tea}}$. We represent the relative camera prediction as $\mathbf{p}_{k,t}=[\mathbf{u}_{k,t},\mathbf{q}_{k,t},\mathbf{f}_{k,t}]$, where $\mathbf{u}$, $\mathbf{q}$, and $\mathbf{f}$ denote translation, unit quaternion, and focal offset, respectively. The pose loss is:
\begin{equation}
\label{eq:pose_loss}
\mathcal{L}_{\mathrm{pose}}
=
\|
\overline{\mathbf{u}}_{k,t}
-
\overline{\mathbf{u}}_{k,t}^{\mathrm{tea}}
\|_1
+
\|
\mathbf{q}_{k,t}
-
\mathbf{q}_{k,t}^{\mathrm{tea}}
\|_1
+
\|
\mathbf{f}_{k,t}
-
\mathbf{f}_{k,t}^{\mathrm{tea}}
\|_1 ,
\end{equation}
where $\overline{\mathbf{u}}$ denotes scale-normalized translation. This decouples relative camera estimation from the metric scale learned by the scale head.

\paragraph{Metric scale loss.}
Following the orthogonal scale learning strategy of LongStream~\citep{cheng2026longstream}, the scale head is supervised independently in log space:
\begin{equation}
\label{eq:scale_loss}
\mathcal{L}_{\mathrm{scale}}
=
\left|
\log s_{k,t}
-
\log s_{k,t}^{\mathrm{gt}}
\right|.
\end{equation}
We apply $\mathcal{L}_{\mathrm{scale}}$ only to training samples with metric-scale supervision; for non-metric datasets, this term is disabled.

\paragraph{Dynamic loss.}
To supervise dynamic scene decomposition, we use SAM3 to generate a pseudo ground-truth mask $M_t\in\{0,1\}^{H\times W}$ for frame $I_t$. Since the Gaussian map is pixel-aligned, the predicted dynamic probability map $P_t\in[0,1]^{H\times W}$ can be directly supervised in image space. We use the binary cross-entropy $\ell_{\mathrm{BCE}}(p,m)=-[m\log p+(1-m)\log(1-p)]$ and compute:
\begin{equation}
\label{eq:dynamic_loss}
\mathcal{L}_{\mathrm{dyn}}
=
\frac{1}{HW}
\sum_{u,v}
\ell_{\mathrm{BCE}}
\left(
P_t(u,v),
M_t(u,v)
\right).
\end{equation}
This supervision encourages the Gaussian head to assign high dynamic probability to time-varying regions and low probability to persistent scene content.

\begin{table*}[t]
\centering
\caption{
Rendering quality comparison on dynamic-scene datasets.
For each scene, metrics are first averaged over 14 uniformly sampled frames,
followed by an equal-weight average across scenes.
Best results are shown in \textbf{bold}, and second-best results are \underline{underlined}.
}
\label{tab:dynamic_rendering}
\scriptsize
\setlength{\tabcolsep}{3pt}
\renewcommand{\arraystretch}{1.05}

\resizebox{\textwidth}{!}{
\begin{tabular}{lccccccccccccccc}
\toprule

& \multicolumn{9}{c}{\textbf{Real-world Dynamic}}
& \multicolumn{6}{c}{\textbf{Synthetic Dynamic}} \\

\cmidrule(lr){2-10}
\cmidrule(lr){11-16}

\textbf{Method}
& \multicolumn{3}{c}{Waymo}
& \multicolumn{3}{c}{SpatialVID}
& \multicolumn{3}{c}{KITTI}
& \multicolumn{3}{c}{PointOdyssey}
& \multicolumn{3}{c}{TartanAir} \\

\cmidrule(lr){2-4}
\cmidrule(lr){5-7}
\cmidrule(lr){8-10}
\cmidrule(lr){11-13}
\cmidrule(lr){14-16}

& PSNR $\uparrow$ & SSIM $\uparrow$ & LPIPS $\downarrow$
& PSNR $\uparrow$ & SSIM $\uparrow$ & LPIPS $\downarrow$
& PSNR $\uparrow$ & SSIM $\uparrow$ & LPIPS $\downarrow$
& PSNR $\uparrow$ & SSIM $\uparrow$ & LPIPS $\downarrow$
& PSNR $\uparrow$ & SSIM $\uparrow$ & LPIPS $\downarrow$ \\

\midrule

AnySplat
& 24.109 & 0.7822 & 0.1998
& \underline{22.096} & \underline{0.7581} & \underline{0.1528}
& 16.003 & 0.4768 & 0.6493
& 20.554 & 0.6911 & 0.3157
& \underline{22.302} & \underline{0.6664} & 0.3025 \\

DGGT
& \underline{24.396} & \underline{0.8125} & \underline{0.1462}
& 21.232 & 0.7350 & 0.1569
& \underline{17.010} & \underline{0.6497} & \underline{0.2911}
& 17.148 & 0.5704 & 0.4264
& 20.537 & 0.6488 & \underline{0.2988} \\

MVSplat
& 18.474 & 0.5291 & 0.3776
& 15.965 & 0.4217 & 0.3917
& 14.845 & 0.3697 & 0.3322
& 17.973 & 0.6093 & 0.4369
& 16.991 & 0.4731 & 0.4868 \\

NeoVerse
& 19.732 & 0.6067 & 0.3596
& 17.400 & 0.5009 & 0.2982
& 13.469 & 0.4817 & 0.4432
& \underline{21.868} & \underline{0.6913} & \underline{0.2834}
& 18.932 & 0.5762 & 0.3259 \\

STORM
& 15.435 & 0.5219 & 0.6682
& 14.606 & 0.3480 & 0.6310
& 11.201 & 0.2619 & 0.7729
& 17.233 & 0.5846 & 0.6094
& 14.562 & 0.3397 & 0.6944 \\

\midrule

\textbf{Ours}
& \textbf{26.777} & \textbf{0.8675} & \textbf{0.0972}
& \textbf{23.994} & \textbf{0.8213} & \textbf{0.1034}
& \textbf{17.708} & \textbf{0.6930} & \textbf{0.2748}
& \textbf{26.787} & \textbf{0.8456} & \textbf{0.1119}
& \textbf{22.348} & \textbf{0.7466} & \textbf{0.1872} \\

\bottomrule
\end{tabular}
}
\end{table*}

\section{Experiments}
\label{sec:experiments}

\subsection{Experimental Setup}
\paragraph{Dataset.} We train our model on SpatialVID\citep{wang2026spatialvid}, Waymo Open Dataset\citep{sun2020waymo}, PointOdyssey\citep{zheng2023pointodyssey}, DL3DV-10K\citep{ling2024dl3dv}, HM3D\citep{ramakrishnan2021hm3d}, Replica\citep{straub2019replica}, ETH3D\citep{schops2017eth3d}, Matterport3D\citep{chang2017matterport3d}, and TartanAir\citep{wang2020tartanair}, covering diverse real and synthetic indoor and outdoor environments with varying scene dynamics, scales, and camera motion patterns. More details about datasets are provided in the supplementary material. 

\paragraph{Implementation Details.} We initialize the geometry transformer, depth DPT head, relative pose head with weights from VGGT\citep{wang2025vggt}, while the remaining layers are initialized randomly. We use AdamW as our optimizer, where $\beta$ = (0.9, 0.95), weight decay 0.05, linear warm-up followed by cosine decay, BF16 mixed precision, and gradient clipping at 0.5. The training loader samples 2–14 context views with probabilities proportional to the squared view count and varies image height in multiples of 16 from 224 to 448. For training losses, we set $\lambda_1=1, \lambda_2 =10, \lambda_3 =1, \lambda_4=0.1, \lambda_5 = 0.05,\lambda_6=0.1$.

\subsection{Quantitative Results}
\label{sec:waymo}

\paragraph{\textbf{Rendering quality.}} We evaluate our method on both outdoor and indoor dynamic-scene benchmarks and compare our method with recent state-of-the-art feed-forward gaussian reconstruction framework. We report PSNR, SSIM and  LPIPS-Alex in this experiment. As is shown in Tab.~\ref{tab:dynamic_rendering}, our method achieve SOTA rendering quality in all metrics.

\begin{figure*}[ht]
\centering

\includegraphics[width=\textwidth]{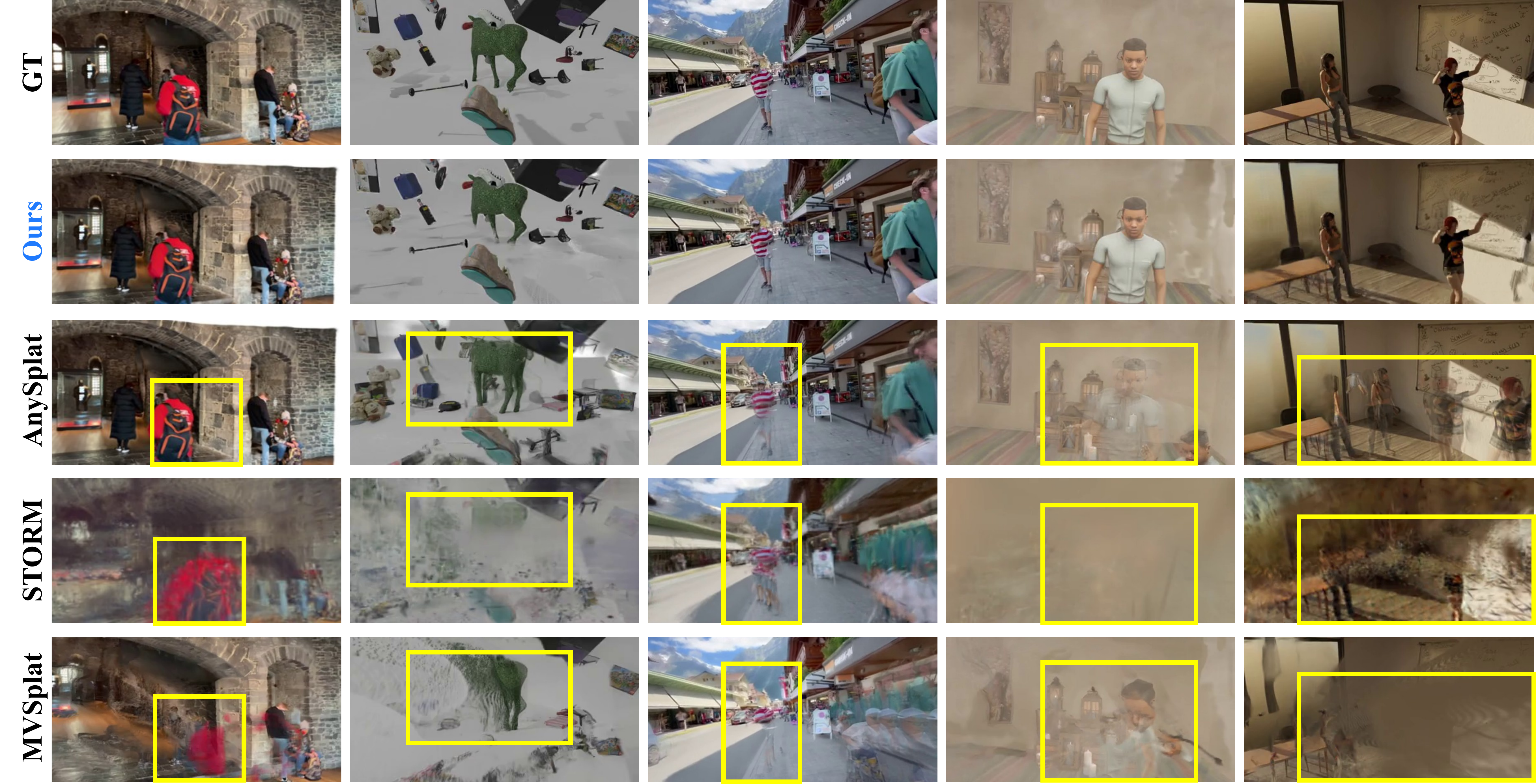}

\caption{Qualitative comparison of novel-view synthesis (NVS) results across diverse scene types. Each column shows a scene, while each row corresponds to the ground truth or a baseline method.}
\label{fig:diverse_qualitative}

\end{figure*}

\begin{table}[t]
\centering
\small
\setlength{\tabcolsep}{3pt}
\caption{Camera tracking performance on the KITTI Odometry benchmark. We evaluate on all 11 sequences comprising 23,201 frames and report sequence-level macro-averaged trajectory errors. Lower is better for all metrics. Best results are shown in \textbf{bold}, and second-best results are \underline{underlined}.}
\label{tab:kitti}
\resizebox{\columnwidth}{!}{
\begin{tabular}{lcccccc}
\toprule
Method
& \shortstack{Sim(3) ATE (m)$\downarrow$}
& \shortstack{SE(3) ATE (m)$\downarrow$}
& \shortstack{Raw ATE (m)$\downarrow$}
& \shortstack{Raw RPE-t (m)$\downarrow$}
& \shortstack{Scale-correctedRPE-t (m) $\downarrow$}
& \shortstack{RPE-r (deg)$\downarrow$} \\
\midrule

STream3R
& 227.226
& \underline{236.056}
& \underline{410.016}
& 1.122
& \underline{11.788}
& \underline{6.451} \\

StreamVGGT
& \underline{205.286}
& 236.072
& 410.059
& \underline{1.088}
& 18.370
& 10.867 \\

VGGT-SLAM
& 219.499
& 416.326
& 582.679
& 342.319
& 39.076
& 9.496 \\

\textbf{Ours}
& \textbf{45.627}
& \textbf{142.064}
& \textbf{283.220}
& \textbf{0.614}
& \textbf{0.255}
& \textbf{0.274} \\

\bottomrule
\end{tabular}
}
\end{table}
\paragraph{\textbf{Camera pose estimation.}} We evaluate all KITTI odometry sequences 00--10 \citep{geiger2012kitti}: 23,201 frames in total, with the longest sequence containing 4,661 frames and covering 5.068\,km. Metrics include Sim(3)- and SE(3)-aligned absolute trajectory error, raw absolute error, and translational/rotational relative error. The results are shown in Tab.\ref{tab:kitti}. Compared to previous online reconstruction method, our method perform better camera pose estimation in long sequence input.

\subsection{Qualitative Results}

\paragraph{Rendering Quality.}
We qualitatively evaluate \ourmethod{} on large-scale driving scenes and diverse indoor and outdoor environments. As shown in Fig.~\ref{fig:waymo_qualitative} and Fig.~\ref{fig:diverse_qualitative}, \ourmethod{} consistently outperforms existing methods in both geometric fidelity and visual quality. Static methods suffer from ghosting and duplicated geometry, while dynamic methods introduce artifacts from inaccurate motion segmentation. In contrast, \ourmethod{} explicitly separates persistent and dynamic content, yielding clean geometry and faithful appearance.

\paragraph{Dynamic Segmentation.}
We further evaluate the predicted dynamic probabilities against SAM3-generated reference masks. Across both Waymo and SpatialVID, \ourmethod{} shows substantially stronger agreement with the reference foreground masks than the compared methods, particularly in foreground coverage and overall overlap. We provide the full protocol, quantitative results, threshold analysis, and fairness considerations in the supplementary material.

\subsection{Ablation Study}
\label{sec:ablation}

We ablate the main components of \ourmethod{} to evaluate their contributions to dynamic-scene reconstruction and long-sequence camera estimation. Specifically, we consider three variants: \emph{w/o dynamic}, which removes dynamic-aware scene decomposition and treats all predicted Gaussians uniformly; \emph{w/o scale head}, which disables metric-scale prediction; and \emph{w/o relative pose}, which removes the proposed keyframe-relative pose formulation. We report rendering quality on Waymo and SpatialVID using PSNR, SSIM, and LPIPS, and evaluate camera trajectories on KITTI and VKITTI2 using Sim(3)- and SE(3)-aligned ATE.

\paragraph{Dynamic-aware reconstruction.}
Table~\ref{tab:ablation_dynamic} evaluates the effect of dynamic-aware scene modeling. Removing the dynamic component substantially degrades rendering quality on both datasets, indicating that explicitly distinguishing time-varying content is important for preventing dynamic observations from being incorrectly accumulated into the persistent scene.

\begin{table}[t]
\centering
\caption{
Ablation of dynamic-aware scene modeling on Waymo and SpatialVID.
Removing the dynamic component consistently degrades rendering quality.
Best results are shown in \textbf{bold}.
}
\label{tab:ablation_dynamic}
\scriptsize
\setlength{\tabcolsep}{5pt}
\renewcommand{\arraystretch}{1.05}

\begin{tabular}{lcccccc}
\toprule
& \multicolumn{3}{c}{Waymo}
& \multicolumn{3}{c}{SpatialVID} \\
\cmidrule(lr){2-4}
\cmidrule(lr){5-7}

Method
& PSNR $\uparrow$
& SSIM $\uparrow$
& LPIPS $\downarrow$
& PSNR $\uparrow$
& SSIM $\uparrow$
& LPIPS $\downarrow$ \\
\midrule

\textbf{\ourmethod{}}
& \textbf{26.777}
& \textbf{0.8675}
& \textbf{0.0972}
& \textbf{23.994}
& \textbf{0.8213}
& \textbf{0.1034} \\

\ourmethod{} w/o dynamic
& 23.011
& 0.7662
& 0.1814
& 21.161
& 0.6529
& 0.1981 \\

\bottomrule
\end{tabular}
\end{table}

\begin{table}[t]
\centering
\caption{
Ablation of relative-pose estimation and metric-scale prediction on KITTI and VKITTI2.
We report Sim(3)- and SE(3)-aligned absolute trajectory error (ATE).
}
\label{tab:ablation_pose}
\scriptsize
\setlength{\tabcolsep}{5pt}
\renewcommand{\arraystretch}{1.05}

\begin{tabular}{lcccc}
\toprule
& \multicolumn{2}{c}{KITTI}
& \multicolumn{2}{c}{VKITTI2} \\
\cmidrule(lr){2-3}
\cmidrule(lr){4-5}

Method
& Sim(3) ATE $\downarrow$
& SE(3) ATE $\downarrow$
& Sim(3) ATE $\downarrow$
& SE(3) ATE $\downarrow$ \\
\midrule

\textbf{\ourmethod{}}
& \textbf{45.63}
& \textbf{142.06}
& \textbf{10.33}
& \textbf{25.05} \\

\ourmethod{} w/o scale head
& 49.66
& 223.80
& 17.76
& 85.00 \\

\ourmethod{} w/o relative pose
& 46.58
& 155.29
& 19.84
& 37.99 \\

\bottomrule
\end{tabular}
\end{table}
\vspace{-2pt}

\paragraph{Relative pose and metric scale.}
Table~\ref{tab:ablation_pose} studies the effect of the relative-pose formulation and metric-scale prediction. Removing the scale head notably increases SE(3)-aligned ATE, especially on VKITTI2, showing that metric calibration is important for maintaining a consistent trajectory scale. Removing relative-pose prediction also degrades trajectory accuracy, confirming the benefit of keyframe-relative camera estimation for long-sequence reconstruction.

\section{Conclusion}
We presented \textbf{\ourmethod{}}, a framework for online 4D Gaussian reconstruction from long, unposed video streams. By incrementally reconstructing local temporal windows and aligning them into a persistent global representation, \ourmethod{} eliminates per-scene optimization while maintaining globally consistent geometry and time-varying content. Extensive experiments demonstrate state-of-the-art reconstruction quality across diverse dynamic scenes. These results establish \ourmethod{} as a scalable approach to persistent online reconstruction of dynamic environments.
\label{maintextend}

\clearpage
\subsection*{AI use statement}
Generative AI tools were used for language polishing and improving the clarity and readability of this manuscript.

\subsection*{Ethics statement}
Reconstruction from driving and indoor imagery can expose people, locations, and private environments. The intended applications are scene understanding and mapping, not personal identification. Data, pretrained checkpoints, and baseline implementations remain subject to their licenses and access conditions; possession of a local copy does not grant redistribution rights. Any public artifact must undergo privacy, licensing, and anonymization review. Reported failures and differences in evaluation protocols are retained to avoid overstating safety, fidelity, or generalization.

\bibliography{references,audit/verified_related}
\bibliographystyle{iclr2027_conference}

\clearpage
\appendix

\newpage
\section*{Supplementary Material}
\label{appendix}

\newcommand{\applabel}{Appendix\xspace}
\renewcommand{\thetable}{\Roman{table}}
\renewcommand{\thefigure}{\Roman{figure}}
\setcounter{table}{0}
\setcounter{figure}{0}

This supplementary material provides additional training details, evaluation protocols, dynamic-segmentation analysis, and limitations omitted from the main paper due to space constraints.

\section{Datasets and Implementation Details}
\label{data}

\paragraph{Training datasets.}
We train \ourmethod{} on SpatialVID~\citep{wang2026spatialvid}, Waymo Open Dataset~\citep{sun2020waymo}, PointOdyssey~\citep{zheng2023pointodyssey}, DL3DV-10K~\citep{ling2024dl3dv}, HM3D~\citep{ramakrishnan2021hm3d}, Replica~\citep{straub2019replica}, ETH3D~\citep{schops2017eth3d}, Matterport3D~\citep{chang2017matterport3d}, and TartanAir~\citep{wang2020tartanair}. These datasets cover real and synthetic indoor and outdoor environments with diverse camera motion, scene scale, and dynamics.

\begin{table}[!ht]
    \centering
    \caption{Overview of the datasets used for training.}
    \label{tab:training_datasets}
    \resizebox{\linewidth}{!}{
    \begin{tabular}{lccccc}
        \toprule
        Dataset & Scene Type & Dynamic & Real / Synthetic & \# Frames & \# Scenes \\
        \midrule
        SpatialVID~\cite{wang2026spatialvid}
            & Indoor / Outdoor & \cmark & Real
            & 7,089 h & 2.7M clips \\
        Waymo~\cite{sun2020waymo}
            & Outdoor & \cmark & Real
            & 390K & 2,030 \\
        PointOdyssey~\cite{zheng2023pointodyssey}
            & Indoor / Outdoor & \cmark & Synthetic
            & $\sim$200K & 159 \\
        DL3DV-10K~\cite{ling2024dl3dv}
            & Indoor / Outdoor & \xmark & Real
            & 51.2M & 10,510 \\
        HM3D~\cite{ramakrishnan2021hm3d}
            & Indoor & \xmark & Real
            & -- & 1,000 \\
        Replica~\cite{straub2019replica}
            & Indoor & \xmark & Real
            & -- & 18 \\
        ETH3D~\cite{schops2017eth3d}
            & Indoor / Outdoor & \xmark & Real
            & 454 & 25 \\
        Matterport3D~\cite{chang2017matterport3d}
            & Indoor & \xmark & Real
            & 194.4K & 90 \\
        TartanAir~\cite{wang2020tartanair}
            & Indoor / Outdoor & \cmark & Synthetic
            & $>$1M & 30 \\
        \bottomrule
    \end{tabular}
    }
\end{table}

\paragraph{Training details.}
We initialize the geometry transformer, DPT depth head, and relative-pose head from VGGT~\citep{wang2025vggt}; the remaining prediction layers are initialized randomly. We use AdamW with $\beta=(0.9,0.95)$ and weight decay $0.05$, followed by linear warm-up and cosine learning-rate decay. Training uses BF16 mixed precision and gradient clipping at $0.5$.

The training loader samples 2--14 input views, with sampling probability proportional to the squared number of views, and varies the image height from 224 to 448 in multiples of 16. The loss weights are $\lambda_1=1$, $\lambda_2=10$, $\lambda_3=1$, $\lambda_4=0.1$, $\lambda_5=0.05$, and $\lambda_6=0.1$. Metric scale supervision is applied only to samples with metric-scale annotations.

\section{Additional Evaluation Protocols}
\label{compare1}

\paragraph{Rendering evaluation.}
We report PSNR, SSIM, and LPIPS-Alex for rendering quality. All methods are evaluated on the same input sequences for each benchmark. No additional image enhancement or post-processing is applied to the rendered outputs.

\paragraph{Camera pose evaluation.}
We evaluate KITTI odometry sequences 00--10~\citep{geiger2012kitti}, containing 23,201 frames in total. We report Sim(3)-aligned ATE, SE(3)-aligned ATE, raw ATE, translational RPE, and rotational RPE. Sim(3) alignment removes global scale, while SE(3) alignment preserves scale error and therefore additionally reflects the quality of metric-scale estimation.

\paragraph{Long-sequence setting.}
All sequences are processed incrementally. The reconstruction network operates on the current bounded chunk rather than jointly processing the complete sequence. Previously reconstructed scene content and camera poses are maintained in the global scene state.

\section{Additional Analysis of Dynamic Segmentation}
\label{sec:supp_dynamic_seg}

The Gaussian head predicts a dynamic probability for every pixel-aligned Gaussian. We further evaluate these predictions against automatically generated SAM3 reference masks.

\paragraph{Reference masks.}
We generate SAM3 masks using the prompts \emph{vehicle}, \emph{person}, and \emph{cyclist}. These masks describe semantic foreground rather than ground-truth physical motion: for example, parked vehicles may be included, while moving vegetation or water may be excluded. Our dynamic head is also trained with automatically generated labels from the same source. Therefore, this experiment measures consistency with a common foreground reference rather than independent zero-shot motion segmentation.

\paragraph{Protocol.}
We evaluate 100 Waymo scenes and 100 SpatialVID scenes. For each scene, 14 frames are uniformly sampled at $224\times448$ resolution, resulting in 1,400 frames per dataset. All methods receive the same RGB inputs, without ground-truth cameras or additional sky masks. Scenes are selected before inference using a fixed random seed and foreground amount, without filtering according to method performance.

For SpatialVID, we use the validation split produced by the current backbone data loader. Since a historical fixed training-scene list is unavailable, we cannot completely exclude possible overlap introduced by earlier preprocessing changes.

\paragraph{Metrics.}
We report IoU, foreground coverage, precision, Dice, and average precision (AP). Coverage is defined as $\mathrm{TP}/(\mathrm{TP}+\mathrm{FN})$, while precision is $\mathrm{TP}/(\mathrm{TP}+\mathrm{FP})$. Metrics are first computed after aggregating the 14 frames of each scene and then averaged across scenes. AP is computed from continuous dynamic scores and does not require selecting a binary threshold.

\begin{table}[t]
\centering
\caption{Consistency with SAM3-generated foreground reference masks.}
\label{tab:dynamic_segmentation}
\resizebox{\linewidth}{!}{
\begin{tabular}{llccccccc}
\toprule
Dataset & Method & Scenes / Frames
& IoU $\uparrow$
& Coverage $\uparrow$
& Precision $\uparrow$
& Dice $\uparrow$
& AP $\uparrow$
& Sec./Frame $\downarrow$ \\
\midrule
Waymo
& \ourmethod{}
& 100 / 1400
& 72.00
& 97.74
& 72.97
& 83.22
& 96.58
& 0.0243 \\
Waymo
& DGGT
& 100 / 1400
& 34.88
& 39.13
& 76.68
& 46.69
& 63.37
& 0.0329 \\
Waymo
& NeoVerse
& 100 / 1400
& 23.02
& 35.40
& 48.15
& 33.09
& 30.51
& 0.0386 \\
\midrule
SpatialVID
& \ourmethod{}
& 100 / 1400
& 58.82
& 95.79
& 59.91
& 72.69
& 91.86
& 0.0251 \\
SpatialVID
& NeoVerse
& 100 / 1400
& 28.25
& 44.85
& 47.72
& 39.16
& 37.77
& 0.0462 \\
SpatialVID
& MoVieS
& 100 / 1400
& 12.40
& 51.48
& 18.68
& 19.70
& 36.04
& 0.0980 \\
\bottomrule
\end{tabular}
}
\end{table}

As shown in Table~\ref{tab:dynamic_segmentation}, \ourmethod{} obtains substantially higher IoU, Dice, and AP on both datasets. Its high foreground coverage is particularly useful for reconstruction, since missing dynamic regions may cause time-varying Gaussians to be incorrectly accumulated into the persistent scene.

\paragraph{Threshold analysis.}
The main reconstruction pipeline uses $p_{\mathrm{dyn}}>0.05$, favoring foreground recall to reduce dynamic false negatives. Since different methods expose different dynamic signals, their native thresholds are not directly comparable. We therefore additionally compare compatible probability outputs using the fixed threshold $0.5$.

\begin{table}[t]
\centering
\caption{Dynamic segmentation using a common probability threshold of $0.5$.}
\label{tab:dynamic_threshold_05}
\begin{tabular}{llccccc}
\toprule
Dataset & Method & Scenes
& IoU $\uparrow$
& Coverage $\uparrow$
& Precision $\uparrow$
& Dice $\uparrow$ \\
\midrule
Waymo
& \ourmethod{}
& 100
& 83.31
& 88.90
& 92.74
& 90.61 \\
Waymo
& DGGT
& 100
& 35.45
& 40.67
& 74.15
& 47.52 \\
\midrule
SpatialVID
& \ourmethod{}
& 100
& 74.93
& 82.46
& 87.28
& 84.32 \\
\bottomrule
\end{tabular}
\end{table}

DGGT uses its native dynamic-logit threshold of $0.5$. NeoVerse uses a velocity threshold of $2\times10^{-4}$ together with cross-view motion verification and therefore cannot be directly compared under a probability threshold. For MoVieS, we estimate the required cameras using an independent official VGGT model from the same RGB inputs and classify points as dynamic when the maximum normalized 3D displacement over the 14 target times exceeds $0.01$.

Runtime includes synchronized model inference and dynamic-mask extraction but excludes model loading, image decoding, export, and metric computation. The additional VGGT preprocessing required by MoVieS is included.

\section{Model Efficiency}
\label{efficiency}

\ourmethod{} achieves a favorable trade-off between reconstruction quality and computational efficiency while remaining scalable to arbitrarily long video streams. As shown in Table~\ref{tab:chunk_runtime_all}, its end-to-end runtime remains manageable as the number of input chunks increases, reaching 315.5 ms per frame even with 100 chunks. In contrast, all competing feed-forward methods fail to process the 100-chunk stream, highlighting their limited scalability to long-term online reconstruction. This scalability stems from our incremental chunk-wise processing, which incorporates new observations without reprocessing previously reconstructed content, enabling \ourmethod{} to operate continuously on unbounded streaming inputs.

\begin{table}[t]
\centering
\caption{
Average end-to-end runtime across all available online chunk experiments.
Runtime is measured as milliseconds per input frame. Lower is better.
}
\label{tab:chunk_runtime_all}
\scriptsize
\setlength{\tabcolsep}{20pt}
\renewcommand{\arraystretch}{1.0}

\resizebox{\columnwidth}{!}{
\begin{tabular}{lcccc}
\toprule

& \multicolumn{4}{c}{End-to-end runtime under different chunk numbers $\downarrow$} \\

\textbf{Method}
& \textbf{2}
& \textbf{3}
& \textbf{5} 
& \textbf{100} \\

\midrule

DGGT
& 300.4 & 308.8 & 405.2 & Failed\\

MVSplat
& 410.2 & 505.2 & 568.6 & Failed\\

NeoVerse
 & 1482.3 & 320.3 & 1313.2 & Failed\\

\textbf{Ours}
& 753.4 & 444.6 & 639.3 & 315.5\\

\bottomrule
\end{tabular}
}
\end{table}

\section{Additional Qualitative Results}
\label{compare2}

We provide additional qualitative results for both driving scenes and diverse indoor and outdoor environments. Consistent with the main paper, static reconstruction methods often produce ghosting or duplicated geometry around moving objects, while imperfect dynamic separation can leave residual artifacts. \ourmethod{} better preserves persistent background geometry while maintaining time-specific foreground content.

\section{Detailed Ablation Studies}
\label{ablations}

We ablate the main components of \ourmethod{}: causal temporal aggregation, keyframe-relative pose prediction, metric scale estimation, and dynamic-aware scene integration.

The causal-attention ablation measures the effect of temporal information exchange under the online constraint. The relative-pose and scale ablations evaluate cross-chunk camera and geometry consistency. Finally, removing dynamic-aware integration tests whether treating all Gaussians as persistent static content introduces artifacts in time-varying regions.

\section{Failure Cases and Limitations}
\label{failure}

\ourmethod{} still has several limitations. First, global camera poses are obtained by propagating relative transformations across chunks, so pose and scale errors may accumulate over very long trajectories without global bundle adjustment or loop closure.

Second, dynamic integration depends on the predicted dynamic probability. False negatives may incorrectly fuse moving content into the persistent scene, while false positives may prevent static geometry from being accumulated correctly. The SAM3-based supervision also emphasizes the selected semantic foreground categories rather than arbitrary physical motion.

Finally, bounded temporal context limits network inference cost but does not bound the size of the accumulated Gaussian scene. Memory consumption can therefore continue to increase when processing very long sequences containing large unexplored regions or extensive dynamic content.

\section{Broader Impacts}
\label{impact}

Online 4D reconstruction can benefit robotics, autonomous systems, simulation, and embodied intelligence by providing continuously updated representations of dynamic environments. However, reconstruction of driving and indoor imagery may contain people, vehicles, locations, or private spaces. The intended use of \ourmethod{} is scene understanding and mapping rather than personal identification. Dataset licenses, privacy requirements, and anonymization procedures should therefore be respected when releasing reconstructed scenes or derived data.

\end{document}